\def\year{2020}\relax
\documentclass[letterpaper]{article} 
\usepackage{aaai20}  
\usepackage{times}  
\usepackage{helvet} 
\usepackage{courier}  
\usepackage[hyphens]{url}  
\usepackage{graphicx} 
\usepackage{graphicx}  
\usepackage{xcolor}
\usepackage{helvet}
\usepackage{courier}
\usepackage{subfigure}
\usepackage{balance}
\usepackage{amsthm}
\usepackage{textcomp,booktabs}
\usepackage{footmisc}
\usepackage[normalem]{ulem}
\usepackage{multirow}
\usepackage{setspace}
\usepackage{bigstrut}
\usepackage{array}
\usepackage{xcolor}
\usepackage{amsfonts}
\usepackage{bm}
\usepackage{url}
\usepackage{threeparttable}

\title{Domain Conditioned Adaptation Network}
\author{Shuang Li,\textsuperscript{\rm 1}
Chi Harold Liu,\textsuperscript{\rm 1}
Qiuxia Lin,\textsuperscript{\rm 1}
Binhui Xie,\textsuperscript{\rm 1}
Zhengming Ding,\textsuperscript{\rm 2}
Gao Huang,\textsuperscript{\rm 3}
Jian Tang\textsuperscript{\rm 4,5}\\
\textsuperscript{\rm 1}School of Computer Science and Technology, Beijing Institute of Technology, China\\
\textsuperscript{\rm 2}Department of Computer, Information and Technology, Indiana University-Purdue University Indianapolis, USA\\
\textsuperscript{\rm 3}Department of Automation, Tsinghua University, China,
\textsuperscript{\rm 4}AI Labs, Didi Chuxing, China\\
\textsuperscript{\rm 5}Department of Electrical Engineering and Computer Science, Syracuse University, USA\\
\{shuangli,chiliu,linqiuxia,binhuixie\}@bit.edu.cn,
zd2@iu.edu,
gaohuang@tsinghua.edu.cn,
jtang02@syr.edu
}

\begin{document}

\maketitle

\begin{abstract}
Tremendous research efforts have been made to thrive deep domain adaptation (DA) by seeking domain-invariant features. Most existing deep DA models only focus on aligning feature representations of task-specific layers across domains while integrating a totally shared convolutional architecture for source and target. However, we argue that such strongly-shared convolutional layers might be harmful for domain-specific feature learning when source and target data distribution differs to a large extent. In this paper, we relax a shared-convnets assumption made by previous DA methods and propose a \textit{Domain Conditioned Adaptation Network (DCAN)}, which aims to excite distinct convolutional channels with a domain conditioned channel attention mechanism. As a result, the critical low-level domain-dependent knowledge could be explored appropriately. As far as we know, this is the first work to explore the domain-wise convolutional channel activation for deep DA networks. Moreover, to effectively align high-level feature distributions across two domains, we further deploy domain conditioned feature correction blocks after task-specific layers, which will explicitly correct the domain discrepancy. Extensive experiments on three cross-domain benchmarks demonstrate the proposed approach outperforms existing methods by a large margin, especially on very tough cross-domain learning tasks.
\end{abstract}

\section{Introduction}

\begin{figure}[tb]
\centering
\includegraphics[width=1.0\columnwidth]{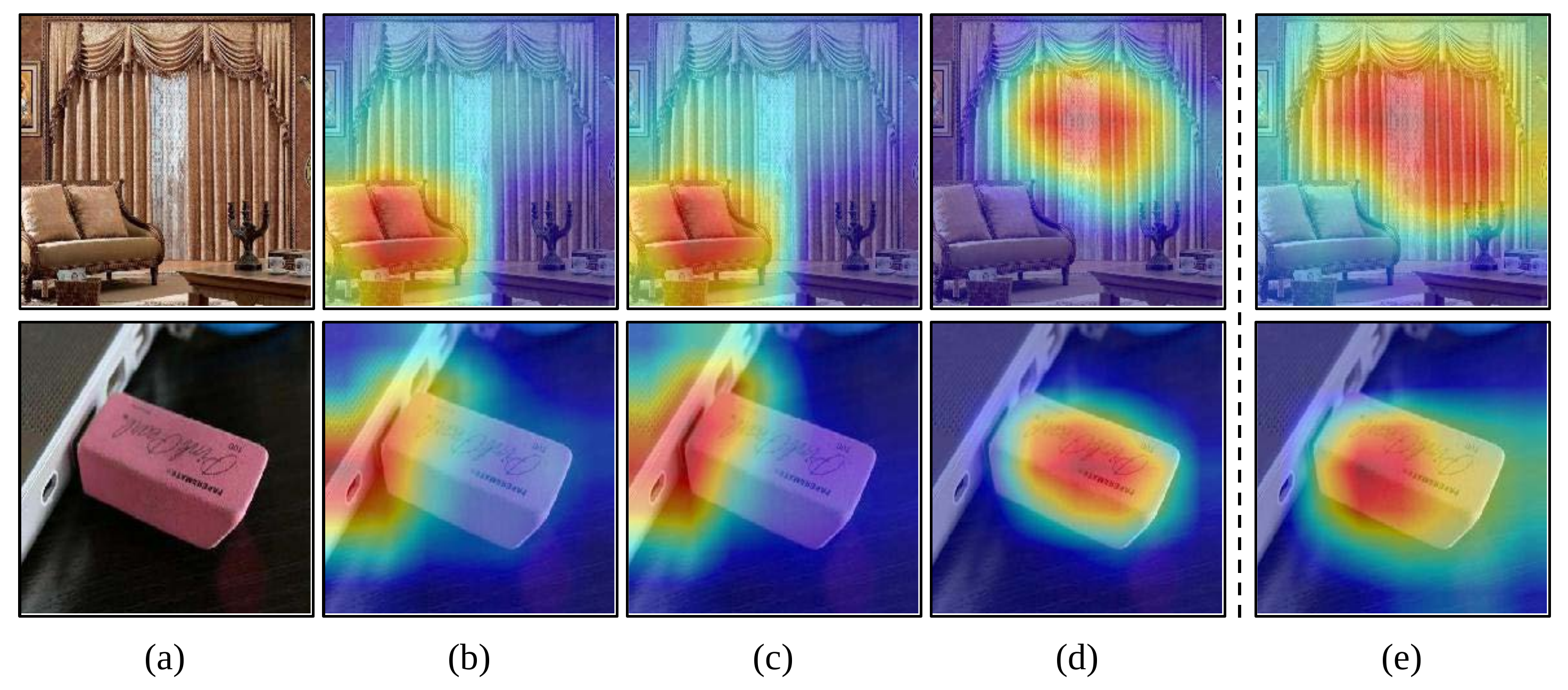}
\caption{Attention visualization of the last convolutional layer of different models on the task Ar$\rightarrow$Rw of Office-Home. Column (a) shows two randomly-chosen target images from classes ``\textbf{curtains}'' and ``\textbf{eraser}''. Column (e) is obtained by the network trained with target ground-truth labels. (b), (c) and (d) represent attention maps of source-only model, DCAN w/o and w/ the proposed domain conditioned channel attention mechanism, respectively. }
\label{Fig_attention_visualization}
\end{figure}

Domain adaptation (DA) \cite{survey} has been witnessed as a promising technique to transfer knowledge from a well-labeled and related source domain to assist the target learning \cite{DIP,DeCAF,DCORAL,openDA}. Previous DA methods generally aim to align domain distributions by either reweighting instances \cite{co-training} or learning domain-invariant features \cite{SCA}. Most recently, deep domain adaptation has been proposed to exploit the powerful feature extraction ability of convolutional neural networks, where they mainly deploy two kinds of strategies to align cross-domain features in the top task-specific fully-connected layers, i.e., discrepancy losses \cite{JAN,MDD,DomainNet,JDDA} and adversarial losses \cite{ADDA,MADA,CDAN,SymNets}. However, they all assume the convolutional layers are universal across different domains in capturing general low-level features based on the analysis of AlexNet \cite{transferable}.

Unfortunately, more challenging cross-domain tasks require much deeper neural networks to achieve promising performance. In this scenario, sharing the totally same convolutional layers in domain adaptation would cause two issues. First, for tough visual cross-domain tasks, there exists a large discrepancy between the visual patterns for the two domains, i.e., Amazon images with clean background and daily life images with complex background. In this case, it is improper to still assume all the filters in each convolutional layer are similarly activated across different domains. When source and target totally share the same convolutional layers, they would fail to capture domain-informative features in low-level stage. Second, only deploying domain discrepancy penalty terms or adversarial losses on the top layers may be less effective, since the gradients of the loss modifications at the task-specific layers will be decayed via back-propagation scheme. As a result, the shared convolutional layers across domains may lose domain-specific knowledge at the start of the very deep convolutional networks.

To verify our judgments, we visualize the attention maps of the last convolutional layer across different models in Figure \ref{Fig_attention_visualization}. Obviously, in column (e), the convolutional layers can accurately capture the most discriminative regions (i.e., curtains and erasers) when trained with target ground-truth labels. However, as shown in column (c), similar to most DA methods, only conducting distribution alignment on the top layers cannot ensure the target model focusing on the desirable regions, while erroneously highlighting the irrelevant objects (i.e., couch and laptop) since the shared-convnets are still affected by the source domain. By contrast, our approach in column (d) performs more similar to (e). This result intuitively manifests that exploring an effective domain-wise convolutional representation learning mechanism is crucial in capturing most important regions for better addressing DA problems.

In this paper, we propose a novel \textit{Domain Conditioned Adaptation Network (DCAN)} for unsupervised visual domain adaptation by exploring domain conditioned channel attention to seek domain-specific knowledge in convolutional layers. Our main idea is to transfer knowledge in two components: domain conditioned channel attention module and task-specific feature correction module. With the introduced domain conditioned channel attention mechanism, we allow large deviation existing in feature representations of different domains. This improves the representation power and flexibility of the network.
Additionally, to enable adapting source discriminative knowledge to target at high-level stage, we plug feature correction block to explicitly learn the domain difference with reference to the target distribution. To sum up, we highlight the three-fold contributions:
\begin{itemize}
    \item First, domain conditioned channel attention module contains two different routes for source and target domains with partially shared parameters. To be specific, channel-wise attention will be jointly learned to activate different channels to adapt source and target data respectively. This flexible module would facilitate extracting more enriched domain-specific knowledge in low-level stage.
    \item Second, we introduce a domain conditioned feature correction block to further explicitly reduce high-level feature distribution discrepancy across domains. Moreover, we design a source-correction regularizer to enhance the effectiveness and stability of this module.
    \item Finally, we provide a comprehensive evaluation of DCAN on three popular cross-domain benchmarks and achieve several state-of-the-art results, especially on the to date largest cross-domain benchmark \textit{DomainNet}. This indicates DCAN can effectively deal with challenging DA problems from small to large scales.
\end{itemize}

\section{Related Work}
Conventional deep DA methods adopt deep neural networks, \textit{e.g.,} AlexNet and ResNet, as the backbone network to seek domain-invariant features in task-specific layer through various statistical moment matching techniques, such as Maximum Mean Discrepancy (MMD) \cite{MMD,DDC}. To name a few, Long \textit{et al.} explore multi-kernel MMD (MK-MMD) metric to minimize marginal distributions of two domains in \cite{DAN}. RTN adapts fused feature by minimizing MMD to jointly learn adaptive classifiers and transferable features \cite{RTN}. Further, Margin Disparity Discrepancy (MDD) is proposed in \cite{MDD} with rigorous generalization bounds. JDDA in \cite{JDDA} aims to match the covariance of source and target with discriminative information preserved. However, all these works enforce source and target data to share one common backbone convolutional network, which usually underestimates the domain mismatch in low-level convolutional stage. Besides, to explore the domain particular knowledge from the network architecture design perspective, domain-specific batch normalization (DSBN) layers are proposed in \cite{DSBN}, which allows source and target data pass through separate batch normalization layers. \cite{DWT-MEC} designs domain-specific whitening transform (DWT) layers after the convolutional layers for the purpose of matching two domains.

An alternative branch of DA is inspired by the Generative Adversarial Networks (GANs) \cite{GAN}, which explores non-discriminative representations by confusing the domain discriminator in a two-player minimax game. In \cite{DA_bp}, a domain adversarial neural network (DANN) is introduced to learn task-specific domain-invariant features. Based on DANN, aiming to alleviate the mode collapse problem, MADA \cite{MADA} trains multiple class-wise domain discriminators according to the number of classes. \cite{CDAN} presents conditional domain adversarial network (CDAN) which conditions the models on the classifier predictive knowledge. By leveraging the formulation in GANs, ADDA proposed in \cite{ADDA} incorporates discriminative modeling, untied weight sharing, and a GAN loss into one framework for unsupervised DA. GTA \cite{GTA} transfers target distribution information to the learned embedding utilizing a generator-discriminator pair. To learn more domain-informative knowledge from lower blocks, \cite{iCAN} presents an incremental collaborative and adversarial network (iCAN). MCD \cite{MCD} introduces a new adversarial paradigm by maximizing the discrepancy between two target classifiers' outputs. Domain-symmetric Networks (SymNets) in \cite{SymNets} construct an additional classifier that shares with source and target classifiers for DA.

Recently, attention mechanism has achieved remarkable performance in various computer vision tasks \cite{Attention}. For DA problems, \cite{DUCDA} proposes an attention transfer process for convolutional domain adaptation with aligning attention maps for two domains. Further, \cite{DAAA} proposes a deep adversarial attention alignment (DAAA) approach to transfer knowledge in all the convolutional layers by attention matching. Considering transferable local and global attentions, TADA in \cite{TADA} aims to highlight transferable regions or images across domains. However, these approaches all focus on space attention knowledge.

Differently, we aim to design a more effective transferable deep model by exploring the adaptation in convolutional layers. To be specific, the domain conditioned channel attention mechanism in convolutional layers would benefit the representation learning of inter-domain features as well as domain-specific ones, making it more flexible and powerful in modeling complex data from different domains.

\section{Domain Conditioned Adaptation Network}\label{sec:method}
\subsection{Preliminary and Motivation}
In unsupervised DA, labeled source domain is expressed as $\mathcal{S} =\{({\boldsymbol{x}_s}_i,{y_s}_i)\}^{n_s}_{i=1}$ of ${n_s}$ samples, where ${\boldsymbol{x}_s}$, $y_s$ denote a source sample and its corresponding label. Similarly, unlabeled target domain is defined as $\mathcal{T} =\{{\boldsymbol{x}_t}_j\}^{n_t}_{j=1}$ of ${n_t}$ samples. Source and target domains have the same $\mathcal{C}_n$ classes, but there exists a domain shift between their feature distributions: $P_s(\boldsymbol{x})\neq P_t(\boldsymbol{x})$. The goal of DA is to learn a classifier that generalizes well on target domain by exploring labeled source data and unlabeled target data in the training stage.

Most existing researches in deep DA are devoted to reducing domain differences only in task-specific layers, assuming the strongly-shared backbone convolutional layers could capture general low-level features across domains. However, we believe these domain alignment methods can only reduce, but not essentially remove the cross-domain discrepancy when domain discrepancy is tremendously large, which is more practical and challenging in real-world applications. Thus, a reasonable consideration is that, both inter-domain features and domain-specific ones should be simultaneously learned in convolutional low-level stage to effectively model complex data from different domains. Furthermore, the cross-domain high-level features should also be explored to facilitate discriminative knowledge transfer. To this end, we propose a novel Domain Conditioned Adaptation Network (DCAN), as illustrated in Figure \ref{Fig_architecture}, to effectively extract domain-specific features with domain conditioned channel attention modules. Meanwhile, the domain mismatch is explicitly minimized with feature correction blocks plugging after the task-specific layers, which are the higher layers of the network.

\subsection{Domain Conditioned Convolutional Learning}
To effectively capture domain-specific information in low-level stage, we propose a domain conditioned channel attention mechanism, which is capable of modeling the independencies between the convolutional channels for source and target respectively. As a result, each domain could perform feature recalibration to extract more domain-specific feature descriptions in convolutional layers. This mechanism will indeed facilitate cross-domain feature alignment in task-specific layers. Moreover, it is straightforward to implement the domain conditioned channel attention module in each residual block of ResNet \cite{resnet}, which is the backbone network of our proposed framework.

\begin{figure}[tb]
\centering
\includegraphics[width=1.0\columnwidth]{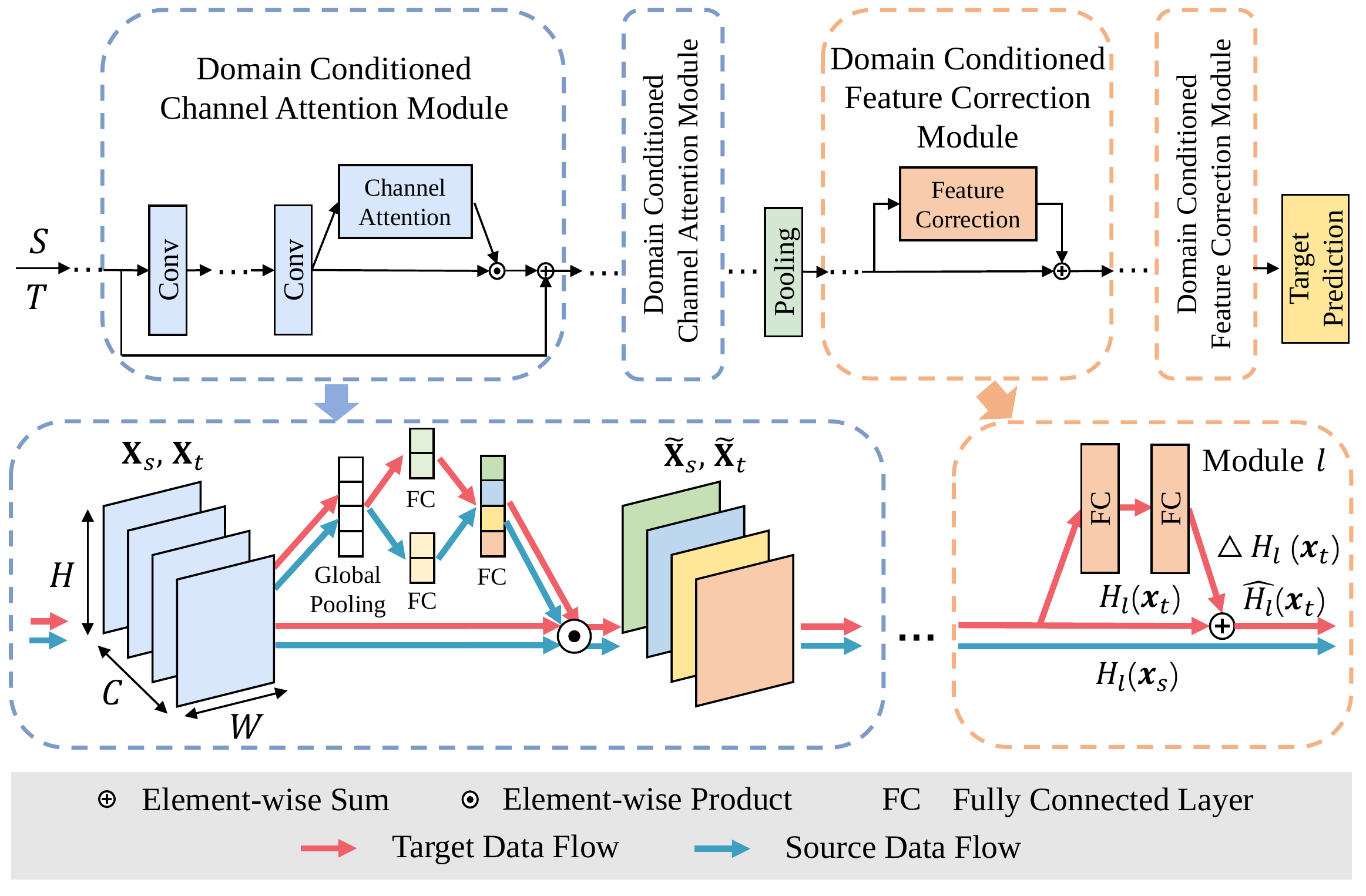}
\caption{Illustration of our proposed Domain Conditioned Adaptation Network (DCAN).}
\label{Fig_architecture}
\end{figure}

Since there exists a certain correlation between source and target, the deep network itself can extract features with generalization abilities. From this respect, source and target data should share most of the network parameters. At the same time, the source and target domains are distributed differently, so the backbone network with supervised source information can extract features that are only sensitive to the source domain samples, which may even bring in negative effect on the feature extraction process of target domain. Hence, we explore the domain conditioned channel attention mechanism, a weakly-shared parameter structure for automatic domain-specific channel selection.

As shown in Figure \ref{Fig_architecture}, we aim to learn a convolutional transformation $F_{conv}:\mathbf{X}_s \rightarrow \widetilde{\mathbf{X}}_s, \mathbf{X}_t \rightarrow \widetilde{\mathbf{X}}_t$, where the source and target input feature tensors are defined as $\mathbf{X}_s=[\mathbf{X}_s^1,...,\mathbf{X}_s^C],\mathbf{X}_t=[\mathbf{X}_t^1,...,\mathbf{X}_t^C] \in \mathbb{R}^{H\times W \times C}$ and the transformed convolutional features are defined as $\widetilde{\mathbf{X}}_s,\widetilde{\mathbf{X}}_t \in \mathbb{R}^{H\times W \times C}$, which consists of $C$ channels with spatial dimensions $H\times W$. The transformation implements an activation of convolutional layers with domain conditioned channel attention, capturing more domain-specific features of images while reducing the influence of useless information.

To be specific, we first utilize global average pooling on each channel to obtain global spatial information of data, by defining it as $\boldsymbol{g} \in 1 \times 1\times C$. Intuitively, such expressive information contains global contextual features for each channel, and generates channel-wise statistics. In order to learn a non-linear interaction across channels and capture channel-wise dependencies for each domain, we assume $\boldsymbol{g}$ passes through different dimensionality-reduction layers with ratio $r$ \footnote{We fix $r=16$ in this paper as \cite{senet}.} into a shape of $1 \times 1\times \frac{C}{r}$, where the upper flow is for target representation $\boldsymbol{f}_t$ and the lower flow is for source representation $\boldsymbol{f}_s$. The benefit of using separate layers for two domains is to assist each domain data in modeling channel dependencies domain-wisely, and learn domain-specific convolutional features automatically. Following with a ReLU operation, $\boldsymbol{f}_s$ and $\boldsymbol{f}_t$ will all pass through the same dimensionality-increasing layer and rescaling function respectively, which resizes the input from $1 \times 1\times \frac{C}{r} $ to $1 \times 1\times C$ as:
\begin{small}
\begin{equation}
\label{channelweight}
  \boldsymbol{v}_s = \sigma\big(\mathbf{W}(\mathrm{ReLU}(\boldsymbol{f}_s))\big),~~
  \boldsymbol{v}_t = \sigma\big(\mathbf{W}(\mathrm{ReLU}(\boldsymbol{f}_t))\big),
\end{equation}
\end{small}
where $\mathbf{W} \in \mathbb{R}^{C\times \frac{C}{r}}$, $\mathrm{ReLU}(\cdot)$ and $\sigma(\cdot)$ refer to the ReLU and Sigmoid functions, respectively. $\boldsymbol{v}_s$ and $\boldsymbol{v}_t$ are the characteristic channel attention vectors with respect to source and target. To achieve the purpose of domain-wise feature selection, the channel attentions will be multiplied to the original feature representations $\mathbf{X}_s$ and $\mathbf{X}_t$ channel-wisely:
\begin{small}
\begin{equation}
\label{selection}
\begin{array}{c}
\widetilde{\mathbf{X}}_s= \boldsymbol{v}_s \odot \mathbf{X}_s = [v_s^1\cdot\mathbf{X}_s^1,...,v_s^C\cdot\mathbf{X}_s^C], \\
  \widetilde{\mathbf{X}}_t= \boldsymbol{v}_t \odot \mathbf{X}_t = [v_t^1\cdot\mathbf{X}_t^1,...,v_t^C\cdot\mathbf{X}_t^C].
  \end{array}
\end{equation}
\end{small}
The domain conditioned channel attention module makes target domain can not only inherit the powerful feature extraction ability from the source network, but also independently learn the importance of each feature channel, which will benefit the recalibration of target domain convolutional features. Meanwhile, the presented domain conditioned channel attention modules only need a small amount of additional parameters and calculations, thus it can be easily applied to most existing deep DA models.

\subsection{Task-specific Feature Alignment via Domain Conditioned Feature Correction}
Recent literature reveals the transferability of features will decrease dramatically along the network \cite{transferable,MDD}, which motivates most state-of-the-art deep DA methods to focus on mitigating the distribution mismatch of high-level features. In this paper, we adapt all the layers corresponding to task-specific features layer-wisely. However, different from previous works, we attempt to explicitly model the feature discrepancy across domains and precisely correct it via the proposed domain conditioned feature correction modules. To be specific, we perform feature adaptation by plugging feature correction block after each task-specific layer.

More specifically, we suppose there are $L$ task-specific layers and perform feature adaptation by plugging feature correction block after each one.
As shown in Figure \ref{Fig_architecture}, for the $l$-th $(l=1,...,L)$ feature correction module, we denote the $l$-th task-specific layer outputs of source data and target data as $H_l(\boldsymbol{x}_s)$ and $H_l(\boldsymbol{x}_t)$, respectively. We enforce target data to pass through not only the original network but also the proposed correction block, consisting of FC, ReLU and FC layers. It is worth noting that, we expect the added correction block $\Delta H_l(\boldsymbol{x}_t)$ to automatically capture the discrepancy between $H_l(\boldsymbol{x}_s)$ and $H_l(\boldsymbol{x}_t)$. Then, the modified target representation can be formalized as $\widehat{H_l}(\boldsymbol{x}_t)=H_l(\boldsymbol{x}_t)+\Delta H_l(\boldsymbol{x}_t)$. Aiming to make $\widehat{H_l}(\boldsymbol{x}_t)$ similar to $H_l(\boldsymbol{x}_s)$, we conduct domain alignment between $\widehat{H_l}(\boldsymbol{x}_t)$ and $H_l(\boldsymbol{x}_s)$ based on the classic MMD criterion \cite{DAN}. The empirical estimation of the discrepancy can be formulated as:
\begin{small}
\begin{equation}\label{f-mmd}
\mathcal{L}_{\mathcal{M}}^l=\bigg\Vert \frac{1}{n_s}\sum_{i=1}^{n_s}\phi \big(H_l({\boldsymbol{x}_s}_i)\big)
-\frac{1}{n_t}\sum_{j=1}^{n_t}\phi\big(\widehat{H_l}({\boldsymbol{x}_t}_j)\big) \bigg\Vert^2_{\mathcal{H}_{\kappa}},
\end{equation}
\end{small}
where $\mathcal{H}_{\kappa}$ is the reproducing kernel Hilbert space (RKHS) with a characteristic kernel $\kappa$, and $\phi$ is the corresponding feature map.
By minimizing (\ref{f-mmd}), we can reduce the distribution difference across domains of the task-specific layers, thereby avoiding feature transferability degradation. Notably, our feature correction block is also added after the softmax layer, which will facilitate transferring category correlation knowledge from source to target in a unified way.

However, this MMD criterion only reduces the domain-wise mismatch such that it may transfer noisy and nonessential information by destroying the structures of source and target. To further avoid the arbitrariness of blocks learning and over-transfer between source and target, we enforce source data to pass through the feature correction blocks for the regularization. Generally, the source domain representation should be unchanged after passing through the feature correction blocks, i.e., the distributions of $H_l(\boldsymbol{x}_s)$ and $\widehat{H_l}(\boldsymbol{x}_s)$ should keep similar. But if we exactly align each class in the source domain, it will result in $\Delta H_l(\boldsymbol{x}_s)\approx 0$. That means the correction blocks will learn nothing for cross-domain feature correction. Thus, we address this problem with a novel regularization loss, which is a random subset of source data to appropriately guide the correction process and enhance the alignment ability of the feature correction blocks. More specifically, we attempt to minimize the MMD metric between each class and a random subset in the source domain. For class $k$, the regularization loss of the $l-$th feature correction module is defined as:
\begin{equation}\scriptsize\label{f-random}
\mathcal{L}_{reg}^{l}=\sum_{k=1}^{\mathcal{C}_n}\bigg\Vert \frac{1}{n_s^{k}}\sum_{{\boldsymbol{x}_s}_i\in\mathcal{S}^{k}}\phi \big(H_l({\boldsymbol{x}_s}_i)\big)
-\frac{1}{|R|}\sum_{{\boldsymbol{x}_s}_j\in R}\phi\big(\widehat{H_l}({{\boldsymbol{x}_s}_j})\big) \bigg\Vert^2_{\mathcal{H}_{\kappa}},
\end{equation}
where $R$ refers to a random subset in the source domain, and $|R|$ is the set size which is stochastic. We define the probability of random sampling for each data is $\frac{p}{\mathcal{C}_n}$ and $p$ is a control factor. To some extent, $\mathcal{L}_{reg}^l$ solves the over-correction problems caused by the added feature correction blocks with the guide of source data. Careful ablation studies investigate the efficacy of key designs of $\mathcal{L}_{\mathcal{M}}^l$ and $\mathcal{L}_{reg}^{l}$.

\renewcommand{\arraystretch}{1.0}
\begin{table*}[!htbp] \footnotesize
  \centering
  \caption{Accuracy (\%) on Office-31 for unsupervised domain adaptation (ResNet-50).}
  \setlength{\tabcolsep}{0.5mm}{
    \begin{tabular}{ccccccccccccccccc|c}
    \toprule
    Methods          &ResNet& JDDA & DAN  & RTN  & DANN & ADDA & MADA & GTA  & MCD            & iCAN           & DAAA           & CDAN           & DSBN           & TADA & SymNets        & MDD            & \textbf{DCAN}   \\
    \hline
    A$\rightarrow$W  & 68.4 & 82.6 & 80.5 & 84.5 & 82.0 & 86.2 & 90.0 & 89.5 & 88.6           & 92.5           & 86.8           & 94.1           & 92.7           & 94.3 & 90.8           & 94.5           & \textbf{95.0}  \\
    D$\rightarrow$W  & 96.7 & 95.2 & 97.1 & 96.8 & 96.9 & 96.2 & 97.4 & 97.9 & 98.5           & 98.8           & \textbf{99.3}  & 98.6           & 99.0           & 98.7 & 98.8           & 98.4           & 97.5           \\
    W$\rightarrow$D  & 99.3 & 99.7 & 99.6 & 99.4 & 99.1 & 98.4 & 99.6 & 99.8 & \textbf{100.0} & \textbf{100.0} & \textbf{100.0} & \textbf{100.0} & \textbf{100.0} & 99.8 & \textbf{100.0} & \textbf{100.0} & \textbf{100.0} \\
    A$\rightarrow$D  & 68.9 & 79.8 & 78.6 & 77.5 & 79.7 & 77.8 & 87.8 & 87.7 & 92.2           & 90.1           & 88.8           & 92.9           & 92.2           & 91.6 & \textbf{93.9}  & 93.5           & 92.6           \\
    D$\rightarrow$A  & 62.5 & 57.4 & 63.6 & 66.2 & 68.2 & 69.5 & 70.3 & 72.8 & 69.5           & 72.1           & 74.3           & 71.0           & 71.7           & 72.9 & 74.6           & 74.6           & \textbf{77.2}  \\
    W$\rightarrow$A  & 60.7 & 66.7 & 62.8 & 64.8 & 67.4 & 68.9 & 66.4 & 71.4 & 69.7           & 69.9           & 73.9           & 69.3           & 74.4           & 73.0 & 72.5           &  72.2          & \textbf{74.9}  \\
    Avg              & 76.1 & 80.2 & 80.4 & 81.6 & 82.2 & 82.9 & 85.2 & 86.5 & 86.5           & 87.2           & 87.2           & 87.7           & 88.3           & 88.4 & 88.4           &  88.9                  &  \textbf{89.5}         \\
    \bottomrule
    \end{tabular}%
    }
  \label{tab:office-31}
\end{table*}

\renewcommand{\arraystretch}{0.85}
\begin{table*}[!htbp] \footnotesize
 \centering
  \caption{Accuracy (\%) on Office-Home for unsupervised domain adaptation (ResNet-50).}
  \setlength{\tabcolsep}{1.0mm}{
    \begin{tabular}{cccccccccccccc}
    \toprule
    Methods &Ar$\rightarrow$Cl &Ar$\rightarrow$Pr &Ar$\rightarrow$Rw &Cl$\rightarrow$Ar &Cl$\rightarrow$Pr &Cl$\rightarrow$Rw &Pr$\rightarrow$Ar &Pr$\rightarrow$Cl &Pr$\rightarrow$Rw &Rw$\rightarrow$Ar &Rw$\rightarrow$Cl &Rw$\rightarrow$Pr &Avg \\
    \midrule
    ResNet & 34.9  & 50.0  & 58.0  & 37.4  & 41.9  & 46.2  & 38.5  & 31.2  & 60.4  & 53.9  & 41.2  & 59.9  & 46.1  \\
    DAN   & 43.6  & 57.0  & 67.9  & 45.8  & 56.5  & 60.4  & 44.0  & 43.6  & 67.7  & 63.1  & 51.5  & 74.3  & 56.3  \\
    DANN & 45.6  & 59.3  & 70.1  & 47.0  & 58.5  & 60.9  & 46.1  & 43.7  & 68.5  & 63.2  & 51.8  & 76.8  & 57.6  \\
    JAN   & 45.9  & 61.2  & 68.9  & 50.4  & 59.7  & 61.0  & 45.8  & 43.4  & 70.3  & 63.9  & 52.4  & 76.8  & 58.3  \\
    DWT & 50.3  & 72.1  & 77.0  & 59.6  & 69.3  & 70.2  & 58.3  & 48.1  & 77.3  & 69.3  & 53.6  & 82.0  & 65.6  \\
    CDAN & 50.7  & 70.6  & 76.0  & 57.6  & 70.0  & 70.0  & 57.4  & 50.9  & 77.3  & 70.9  & 56.7  & 81.6  & 65.8  \\
    TADA  & 53.1  & 72.3  & 77.2  & 59.1  & 71.2  & 72.1  & 59.7  & 53.1  & 78.4  & 72.4  & 60.0  & 82.9  & 67.6  \\
    SymNets & 47.7  & 72.9  & 78.5  & 64.2  & 71.3  & 74.2  & 64.2  & 48.8  & 79.5  & \textbf{74.5} & 52.6  & 82.7  & 67.6  \\
    MDD   & \textbf{54.9} & 73.7  & 77.8  & 60.0  & 71.4  & 71.8  & 61.2  & \textbf{53.6} & 78.1  & 72.5  & \textbf{60.2} & 82.3  & 68.1  \\
    \midrule
    \textbf{DCAN}  & 54.5  & \textbf{75.7} & \textbf{81.2} & \textbf{67.4} & \textbf{74.0} & \textbf{76.3} & \textbf{67.4} & 52.7  & \textbf{80.6} & 74.1  & 59.1  & \textbf{83.5} & \textbf{70.5} \\
    \bottomrule
    \end{tabular}%
    }
  \label{tab:office-home}
\end{table*}%

\subsection{Overall Formulation of DCAN}
For unsupervised domain adaptation, we aim to jointly seek a classifier after we couple source and target domains. Since only source data are well-labeled, we can build a source classifier by minimizing the loss function as:
\begin{equation}\label{f-source}\small
\min_{G}~~~~\mathcal{L}_{s}=\frac{1}{n_s}\sum^{n_s}_{i=1}\mathcal{E}(G({\boldsymbol{x}_s}_i),{y_s}_i),
\end{equation}
where $\mathcal{E}(\cdot,\cdot)$ is the cross-entropy loss function and $G(\cdot)$ is the learned predictive model. However, such a loss only learns a source sensitive representation mapping, and therefore the source classifier may not generalize well on target domain due to the existing domain distribution bias. Note that our target domain is unlabeled, it is reasonable to exploit the entropy minimization principle \cite{entropy} as \cite{SymNets,RTN} to increase the discrimination of the learned models. If we define the $k-$th class-conditional probability of target data ${\boldsymbol{x}_t}$ predicted by $G(\cdot)$ as $G^{(k)}({\boldsymbol{x}_t})$, the target entropy loss can be computed as:
\begin{equation}\label{entropy}\small
\min_{G}~~~~\mathcal{L}_e=-\frac{1}{n_t}\sum_{j=1}^{n_t}\sum_{k=1}^{\mathcal{C}_n}G^{(k)}({\boldsymbol{x}_t}_j)
\mathrm{log}G^{(k)}({\boldsymbol{x}_t}_j).
\end{equation}

To this end, we integrate all the components and obtain the following overall objective of DCAN as:
\begin{equation}\label{objective}\small
\min_{G}~~~~\mathcal{L}=\mathcal{L}_s+\alpha\sum_{l=1}^{L}(\mathcal{L}_\mathcal{M}^l+\mathcal{L}_{reg}^l)+\beta\mathcal{L}_e,
\end{equation}
where $\mathcal{L}_s$, $\mathcal{L}_e$ are the source classification and target entropy losses. $\mathcal{L}_\mathcal{M}^l$ and $\mathcal{L}_{reg}^l$ represent the task-specific feature alignment and regularization losses for the $l-$th domain conditioned feature correction module. $\alpha$ and $\beta$ are two positive trade-off parameters.

\renewcommand{\arraystretch}{0.65}
\begin{table*}[!htbp] \footnotesize
  \centering
  \caption{Accuracy (\%) on DomainNet for unsupervised domain adaptation. ($\dagger$ Implement according to source code.)}
  \setlength{\tabcolsep}{0.4mm}{
  \begin{threeparttable}
    \begin{tabular}{ccccccccccccccc}
    \toprule
    Networks & Methods & inf$\rightarrow$qdr & inf$\rightarrow$rel & inf$\rightarrow$skt & qdr$\rightarrow$inf & qdr$\rightarrow$rel & qdr$\rightarrow$skt & rel$\rightarrow$inf & rel$\rightarrow$qdr & rel$\rightarrow$skt & skt$\rightarrow$inf & skt$\rightarrow$qdr & skt$\rightarrow$rel & Avg \\
   \midrule
    & ResNet & 2.3   & 40.6  & 20.8  & 1.5   & 5.6   & 5.7   & 17.0 & 3.6 & 26.2 & 11.3 & 3.4 & 38.6 & 14.7 \\
    & MCD$^{\dagger}$   & 1.6 & 35.2 & 19.7 & 2.1 & 7.9 & 7.1 & 17.8 & 1.5 & 25.2 & 12.6 & 4.1 & 34.5 & 14.1 \\
    ResNet-50& CDAN$^{\dagger}$  & 1.9   & 36.3  & 21.3  & 1.2 & 9.4 & 9.5 & \textbf{18.3} & 3.4 & 24.6  & 14.7 & 7.0 & 36.6 & 15.4  \\
    & MDD$^{\dagger}$  & 3.6   & 40.0  & 19.2  & 2.0   & 9.2  & 7.7  & 16.8  & \textbf{4.5}   & 27.7  & 14.0  & 7.4 & 42.0 & 16.2  \\
    \cmidrule{2-15}& \textbf{DCAN}  & \textbf{3.8} & \textbf{51.1} & \textbf{24.4} & \textbf{3.7} & \textbf{14.5} & \textbf{12.3}  & 17.5  & 2.7 & \textbf{31.1}  & \textbf{16.2}  & \textbf{8.7}   & \textbf{53.2}  & \textbf{19.9}  \\
   \midrule
    & ResNet & 3.6   & 44.0  & 27.9  & 0.9   & 4.1   & 8.3   & \textbf{22.2}  & 6.4   & \textbf{38.8}  & 15.4  & \textbf{10.9}  & 47.0  & 19.1  \\
    ResNet-101& ADDA  & 3.2   & 26.9  & 14.6  & 2.6   & 9.9   & 11.9  & 14.5  & \textbf{12.1}  & 25.7  & 8.9   & 14.9  & 37.6  & 15.2  \\
    & MCD   & 1.5   & 36.7  & 18.0  & 3.0   & 11.5  & 10.2  & 19.6  & 2.2   & 29.3  & 13.7  & 3.8   & 34.8  & 15.4  \\
    \cmidrule{2-15}& \textbf{DCAN} & \textbf{5.9} & \textbf{54.6} & \textbf{28.5} & \textbf{5.6} & \textbf{18.4}  & \textbf{16.2}  & 18.5  & 4.0   & 33.4  & \textbf{17.3}  & 10.1  & \textbf{55.3} & \textbf{22.3} \\
   \midrule
    & ResNet & 4.7   & 45.5  & 29.6  & 1.8   & 6.3   & 9.4   & \textbf{24.4}  & 6.2   & \textbf{39.9}  & 18.2  & 12.5  & 47.4  & 20.5  \\
    ResNet-152& SE & 1.2   & 13.1  & 6.9   & 3.9   & 16.4  & 11.5  & 12.9   & 3.7   & 26.3  & 7.8   & 7.7   & 28.9  & 11.7  \\
    \cmidrule{2-15}& \textbf{DCAN}  & \textbf{8.8} & \textbf{54.2} & \textbf{31.7} & \textbf{5.6} & \textbf{20.6} & \textbf{17.1} & 21.9 & \textbf{8.0} & 37.3 & \textbf{19.5} & \textbf{16.5} & \textbf{56.8} & \textbf{24.8} \\
    \bottomrule
    \end{tabular}

    \end{threeparttable}
    }
  \label{tab:domainnet}
\end{table*}%

\section{Experiment}\label{sec:experiment}
\subsection{Experimental Setup}
\noindent\textbf{Office-31} \cite{Office31} is a popular object dataset with 4110 images and 31 classes under office settings. It consists of three distinct domains: Amazon (\textbf{A}), Webcam (\textbf{W}) and DSLR (\textbf{D}). As \cite{MDD}, we construct 6 cross-domain tasks: \textbf{A}$\rightarrow$\textbf{W}, ..., \textbf{D}$\rightarrow$\textbf{W}.

\noindent\textbf{Office-Home} \cite{Office-Home} is a challenging benchmark with totally 15588 images, containing 65 classes from 4 domains: Artistic images (\textbf{Ar}), Clip Art (\textbf{Cl}), Product images (\textbf{Pr}) and Real-World images (\textbf{Rw}). And we build 12 adaptation tasks: \textbf{Ar}$\rightarrow$\textbf{Cl}, ..., \textbf{Rw}$\rightarrow$\textbf{Pr}.

\noindent\textbf{DomainNet} is the largest visual domain adaptation dataset so far, and involves about 0.6 million images with 345 categories that evenly spread in 6 domains: Clipart (\textbf{clp}), Infograph (\textbf{inf}), Painting (\textbf{pnt}), Quickdraw (\textbf{qdr}), Real (\textbf{rel}), Sketch (\textbf{skt}). We use all released 4 domains with total 471,414 images: \textbf{inf} (53,201), \textbf{qdr} (172,500), \textbf{rel} (175,327) and \textbf{skt} (70,386) to build 12 adaptation tasks.
Following \cite{DomainNet}, each domain is split into training and test sets. Only training sets of both domains are involved in the training procedure, and the results of target test set are reported.

We implement our approach using PyTorch, and use ResNet \cite{resnet} as backbone networks.
In the experiments, we use a small batch of 32 samples per domain, therefore we freeze the BN layers and only update the weights of other layers through back-propagation.
Besides, we set the learning rate of the classifier layer to be 10 times that of the other layers, while the domain conditioned feature correction blocks are 1/10 times because of its precision.
We follow the standard evaluation protocols for unsupervised domain adaptation, in which source data are all labeled while target data are unlabeled.
All the images are cropped to 224 $\times$ 224 and each domain transfer task is evaluated by averaging three random experiments.
We adopt stochastic gradient descent (SGD) with momentum of 0.9 and the learning rate strategy as described in \cite{DA_bp}.
Moreover, we use the importance weighted cross-validation method as \cite{MDD} to select hyper-parameters.
The values of coefficient $\alpha$, $\beta$ are fixed to 1.5 and 0.1, $p$ is 0.8 chosed from $\{0.2, 0.4, 0.6, 0.8, 1\}$.

\subsection{Comparison Results}
\noindent\textbf{Compared Approaches:}
To better illustrate the effectiveness of our method, we take several state-of-the-art deep domain adaptation methods as baselines, including DAN \cite{DAN}, DANN \cite{DA_bp}, RTN \cite{RTN}, ADDA \cite{ADDA}, JAN \cite{JAN}, MADA \cite{MADA}, SE \cite{SE}, MCD \cite{MCD}, iCAN \cite{iCAN}, GTA \cite{GTA}, CDAN \cite{CDAN}, DAAA \cite{DAAA}, JDDA \cite{JDDA}, DSBN \cite{DSBN}, DWT \cite{DWT-MEC}, TADA \cite{TADA}, SymNets \cite{SymNets} and MDD \cite{MDD}.
Note that partial reported results are copied from their corresponding papers if the experiment setup is the same.

\noindent\textbf{Experiment Results on Office-31:}
We summarize the results in Table \ref{tab:office-31}.
It is desirable that our DCAN dramatically overpasses all recent methods on hard tasks in which the difference between the domains is significant and still attains comparable results on easy tasks.
More specifically, the accuracies of DCAN are \textbf{2.6\%} and \textbf{0.5\%} higher than MDD and DSBN, the second best methods, under these hard tasks: D$\rightarrow$A and W$\rightarrow$A. Although DAAA and SymNets win the first place in task D$\rightarrow$W and A$\rightarrow$D, respectively, the slightly lower results of DCAN are because these two tasks have over 90\% accuracies. In other words, if two domains are much similar, our domain-specific feature learning may not further enhance the performance.
Despite this, DCAN still owns the best performance in average accuracy.

\noindent\textbf{Experiment Results on Office-Home:}
As reported in Table \ref{tab:office-home}, we can notice that DCAN obtains significant improvements over previous methods, highly affirming the effectiveness of the proposed domain conditioned channel attention mechanism and feature correction block
do learn more domain-invariant representations. Therefore, it can achieve more convincing results on this challenging datset.
Overall, DCAN gets the highest average accuracy of \textbf{70.5\%} among all the compared methods and in particular establishes new state-of-the-art records for Office-Home dataset.

\noindent\textbf{Experiment Results on DomainNet:}
In Table \ref{tab:domainnet}, we compare DCAN with previous approaches based on different backbone networks on DomainNet.
On average, our DCAN obtains  an improvement of \textbf{5.2\%}, \textbf{3.2\%} and \textbf{4.3\%} over the baselines using ResNet-50/101/152 respectively.
It is noteworthy that MCD based on ResNet-50 or ResNet-101 performs worse than their source only models, which manifests negative transfer \cite{survey} phenomena occurs.
Similar trends can be found in ADDA and SE approaches.
We believe that the challenge of DomainNet is ascribed to its large class size, which greatly increases the difficulty of tasks. Although ResNet-101\&152 based DCAN have more degradation cases than ResNet-50 based DCAN, the accuracy increases gradually when deepening network. This indicates deeper networks can mitigate domain mismatch better. Besides, most degradations occur in tasks when source is ``\textbf{rel}'', as the complex information in source may misguide target discriminative feature learning when conducting alignment forcefully. However, DCAN still performs better in most cases, which substantiates our method is suitable to very large scale domain adaptation.
Ultimately, we draw a conclusion that DCAN can capture enriched information to help learn more transferable features.
\begin{table}[tb] \scriptsize
  \centering
  \caption{Ablation Study}
  \setlength{\tabcolsep}{0.5mm}{
    \begin{tabular}{cccccccc}
    \toprule
    Methods      & A$\rightarrow$W  & D$\rightarrow$W  & W$\rightarrow$D  & A$\rightarrow$D  & D$\rightarrow$A  & W$\rightarrow$A  & Avg \\
    \midrule
    ResNet-50     & 68.4 & 96.7 & 99.3 & 68.9 & 62.5 & 60.7 & 76.1 \\
    DCAN (w/o $\mathcal{L}_\mathcal{M}^1+\mathcal{L}_{reg}^1$) & 90.1 & 96.5 & 100.0 & 86.5 & 63.6 & 69.6 & 84.4 \\
    DCAN (w/o $\mathcal{L}_\mathcal{M}^2+\mathcal{L}_{reg}^2$) & 93.2 & 97.4 & 100.0 & 94.4 & 71.5 & 70.0 & 87.8 \\
    DCAN (w/o $\mathcal{L}_{reg}^1$)  & 93.8 & 97.6 & 100.0 & 90.6 & 72.5 & 71.8 & 87.7 \\
    DCAN (w/o $\mathcal{L}_{reg}^2$)  & 93.8 & 97.1 & 100.0 & 91.8 & 75.4 & 73.4 & 88.6 \\
    DCAN (w/o $\mathcal{L}_e$)   & 92.1 & 97.1 & 100.0  & 91.0 & 75.1 & 75.3 & 88.4       \\
    DCAN (w/o CA)  & 93.3 & 97.9 & 100.0 & 91.8 & 72.2 & 69.9 & 87.5 \\
    \textbf{DCAN}  & 95.0 & 97.5 & 100.0 & 92.6 & 77.2 & 74.9 & \textbf{89.5} \\
    \bottomrule
    \end{tabular}
   }
  \label{tab:ablation}
\end{table}%

\subsection{Empirical Analysis}
\noindent\textbf{Ablation Study:}
In this section, we conduct thorough analysis to investigate the efficacy of key designs of our proposed DCAN on the Office-31 dataset.
First, based on ResNet-50, we have two domain conditioned feature correction blocks: one is after the pooling layer, and another is after the softmax layer.
We respectively remove the task-specific feature correction loss $\mathcal{L}_\mathcal{M}^1+\mathcal{L}_{reg}^1$/$\mathcal{L}_\mathcal{M}^2+\mathcal{L}_{reg}^2$ and regularization loss $\mathcal{L}_{reg}^1$/$\mathcal{L}_{reg}^2$ for the two modules from overall objective (\ref{objective}),
which are denoted as ``DCAN (w/o $\mathcal{L}_\mathcal{M}^1+\mathcal{L}_{reg}^1$)/(w/o $\mathcal{L}_\mathcal{M}^2+\mathcal{L}_{reg}^2$)" and ``DCAN (w/o $\mathcal{L}_{reg}^1$)/(w/o $\mathcal{L}_{reg}^2$)".
Also, the exclusion of target entropy loss $\mathcal{L}_e$ is denoted as ``DCAN (w/o $\mathcal{L}_e$)".
Besides, to explore the effects of our proposed domain conditioned channel attention mechanism, we adapt the shared-convnets similar to most DA methods, which is denoted as ``DCAN (w/o CA)".

The results are shown in Table \ref{tab:ablation}, it is clear that full method outperforms other variants and achieves large improvements.
``DCAN (w/o $\mathcal{L}_{reg}^1$)/(w/o $\mathcal{L}_{reg}^2$)" is inferior to DCAN with an average decrease of \textbf{1.4\%}, while performing better than ``DCAN (w/o $\mathcal{L}_\mathcal{M}^1+\mathcal{L}_{reg}^1$)/(w/o $\mathcal{L}_\mathcal{M}^2+\mathcal{L}_{reg}^2$)", verifying the effectiveness of regularization loss and feature correction blocks in the feature alignment.
DCAN enhances the performance over ``DCAN (w/o $\mathcal{L}_e$)" , testifying usefulness of entropy minimization principle for DA.
It is can be observed that the adaptation performance of ``DCAN (w/o CA)" suffers a degradation of \textbf{2.0\%}, manifesting the importance of the proposed channel attention mechanism to explore the critical low-level domain-dependent knowledge.

\begin{figure}[tb]
\centering
\includegraphics[width=\columnwidth]{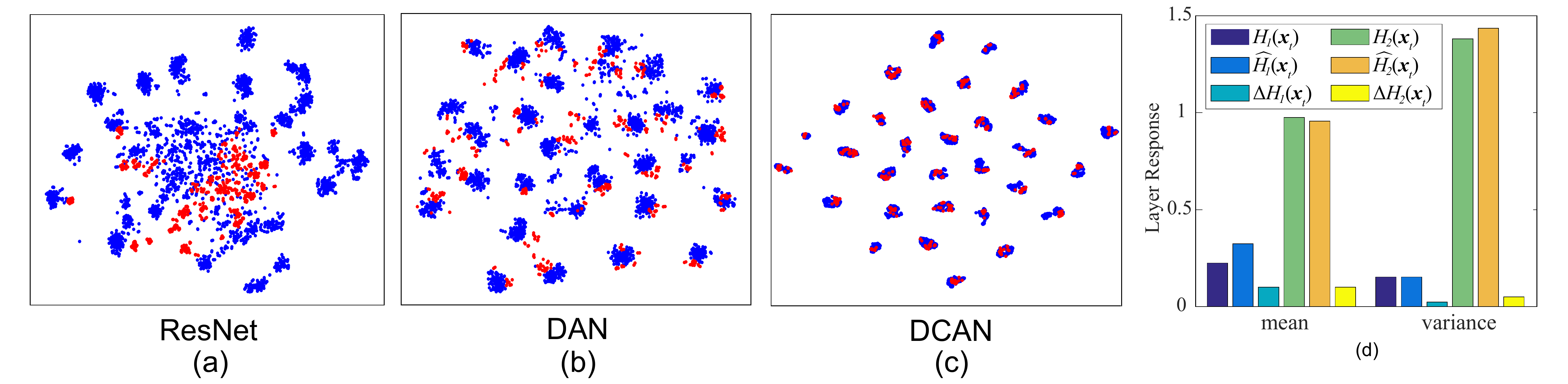}
\caption{(Best viewed in color.) The t-SNE visualizations of (a) ResNet, (b) DAN and (c) DCAN on task \textbf{A}$\rightarrow$\textbf{W} of Office-31, where blue points are source domain data and red points are target domain data.}
\label{Fig_tsne}
\end{figure}

\noindent\textbf{Feature Visualization:}
Figure \ref{Fig_tsne} show the t-SNE \cite{tsne} embedding of feature representations learned by several methods, in which each category is represented as a cluster and domains have different colors.
In ResNet-50, the source and target domains are totally mismatched. In DAN, categories are not aligned very well between domains. However, DCAN shows greater ability of making inter-class separated and intra-class clustered.

\begin{figure}[tb]
\centering
\subfigure[Attention Value Difference]{
    \includegraphics[width=0.95\columnwidth]{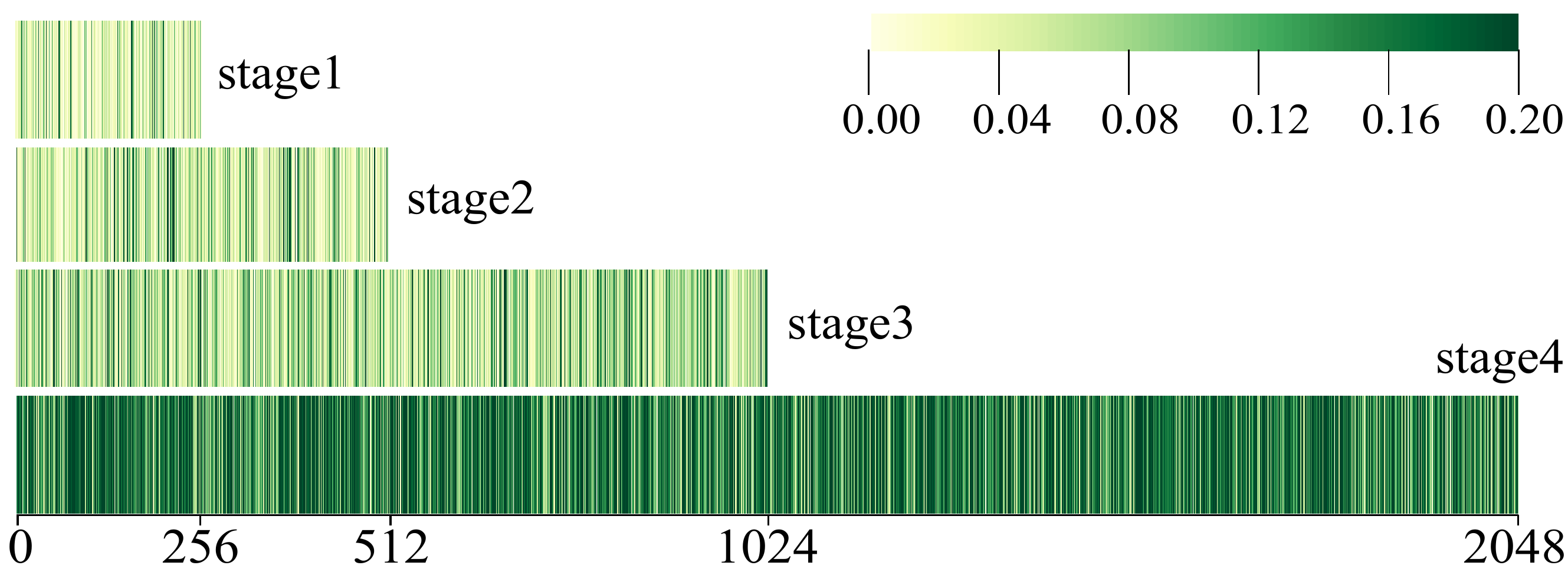}
    \label{Fig_attention1}
}
\subfigure[Attention Difference Comparison]{
    \includegraphics[width=0.95\columnwidth]{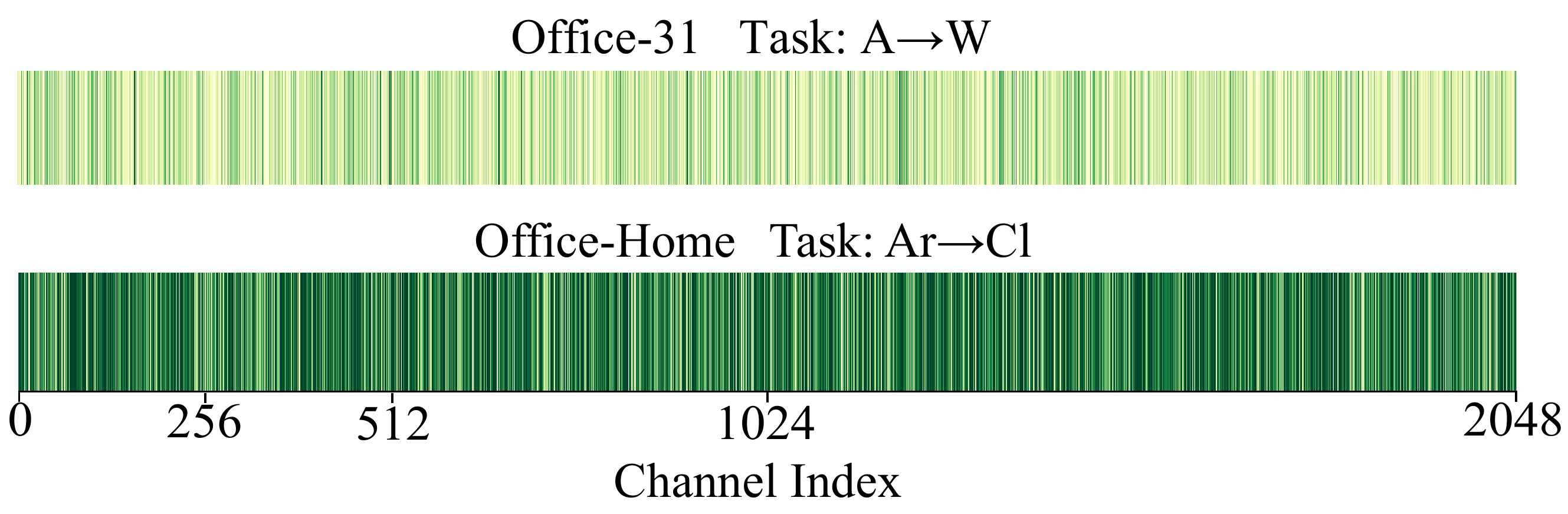}
    \label{Fig_attention2}
}
\caption{(a) The heat-map of attention value difference between source and target in a trained DCAN on task \textbf{Ar}$\rightarrow$\textbf{Cl} (Office-Home). The color of each vertical line represents the degree of attention difference across domains; (b) Attention difference comparison between task \textbf{A}$\rightarrow$\textbf{W} (Office-31) and task \textbf{Ar}$\rightarrow$\textbf{Cl} (Office-Home) at stage4.}
\label{Fig_attention}
\end{figure}
\noindent\textbf{Domain Conditioned Channel Attention Analysis:}
Intuitively, the proposed domain conditioned channel attention mechanism can facilitate learning domain-specific convolutional features by increasing the sensitivity to informative channels and suppressing the useless ones. To provide a clearer picture about the behaviour, we study the channel attention values of source and target domains.

In ResNet-50, the convolutional layers gradually increase the channel size of the input images from 3 to 2048. We compute the average attention values for all the source and target samples in the last domain conditioned channel attention module in each stage (immediately prior to downsampling). We refer to each stage as stage 1, 2, 3 and 4 with the channel numbers as 256, 512, 1024 and 2048, respectively. Specifically, in each stage, we denote convolutional channel attention values of source and target as $\overline{\boldsymbol{v}_s}$ and $\overline{\boldsymbol{v}_t}$. Then for the $m$-th channel, the value of $|\overline{v^m_s}-\overline{v^m_t}|$ indicates the excitation difference of the channel from source and target.

Figure \ref{Fig_attention1} shows the heat-map of attention difference between source and target in each stage. The darker the color, the larger the attention value difference across domains. We observe that the attention difference of the first three stages is obviously smaller than that of the stage 4, since the color of the first three stages are much brighter. Meanwhile, as the number of channels increases, the color becomes darker. The above observations indicate that the source and target networks extract more general low-level features in the early stages of convolutional layers, while they extract more domain-specific features in the later stages of convolutional layers. This observation is similar to the conclusion in \cite{transferable}, and could provide us new insights to design more powerful deep convolutional structure for DA. Therefore, completely sharing convolutional layers across two domains might be improper, which verifies our argument that we should build weakly-shared convolutional structures.

Similar to the previous calculations, Figure \ref{Fig_attention2} illustrates the attention difference comparison between task \textbf{A}$\rightarrow$\textbf{W} of Office-31 and task \textbf{Ar}$\rightarrow$\textbf{Cl} of Office-Home at stage4. Clearly, the figure in the bottom is more darker than the upper one, which indicates larger channel attention value difference for the harder task \textbf{Ar}$\rightarrow$\textbf{Cl}. This further proves our statement that convolutional features should be more domain-specific due to larger domain discrepancy.

\section{Conclusion}
In this paper, we presented a Domain Conditioned Adaptation Network (DCAN) to simultaneously learn domain-specific features in convolutional stage and effectively mitigate the domain mismatch in task-specific layers.
Our designed domain conditioned channel attention module could enrich domain-specific knowledge in low-level stage so as to facilitate subsequent feature migration.
As for high-level feature alignment, we explored feature correction blocks to align marginal and output distributions across domains. With uncomplicated loss function, each of the components could be easily inserted into any layer of original network.
Experiment results demonstrated that DCAN achieved better performance compared with other deep DA methods, especially when it came to very tough cross-domain tasks.

\section{Acknowledgments}
This work was supported by the National Natural Science Foundation of China (61902028).

\end{document}